# Marking-up multiple views of a Text: Discourse and Reference

**Dan Cristea**
Department of Computer Science
University "A.I.Cuza" Iasi
Iasi, RO-6600 Romania
dcristea@infoiasi.ro

**Nancy Ide**
Department of Computer Science
Vassar College
Poughkeepsie, NY, USA
ide@cs.vassar.edu

**Laurent Romary**
Loria-CNRS
B.P. 239
F-54506 Vandoeuvre Lès Nancy
romary@loria.fr

**Abstract**

We describe an encoding scheme for discourse structure and reference, based on the TEI Guidelines and the recommendations of the Corpus Encoding Specification (CES). A central feature of the scheme is a CES-based data architecture enabling the encoding of and access to multiple views of a marked-up document. We describe a tool architecture that supports the encoding scheme, and then show how we have used the encoding scheme and the tools to perform a discourse analytic task in support of a model of global discourse cohesion called *Veins Theory* (Cristea & Ide, 1998).

## 1. Introduction

Recent work on discourse processing has demonstrated the need for large corpora annotated for relational structures in discourse (Cristea and Webber, 1997; Marcu, 1997). Although corpora marked for discourse structure are beginning to exist,[1] they are typically marked using *ad hoc* encoding formats that are designed to accommodate a specific piece of software and/or research need. No coherent, consistent and, above all, standardized encoding scheme for discourse structure currently exists, and as a result, it is common that available corpora require considerable effort to be generally usable for discourse study.

We have taken a more principled approach to the development of an encoding scheme for discourse structure annotation. Our work grows out of our own need for corpora annotated for discourse structure and reference. We describe elsewhere Cristea & Ide (1998) an approach to long-distance reference resolution that demonstrates the relation between discourse cohesion and coherence and discourse structure, called *Veins Theory (VT)*. VT is centered around the identification of "veins" over discourse structure trees such as those defined in Rhetorical Structure Theory (RST) (Mann and Thompson, 1987). To validate our theory, it is necessary to test it on a large sample of real data that is annotated for discourse structure and reference. However, no existing scheme currently supports this kind of markup to the extent required for our work. Therefore, we devised an encoding scheme that provides for reference annotation *and* allows for encoding discourse structure, which both eliminates interference between the two encodings and supports automatic extension.

In this paper, we describe our annotation scheme, realized in an SGML/XML[2] format compatible with the Text Encoding Initiative (TEI) Guidelines (Sperberg-McQueen & Burnard, 1994) and the Corpus Encoding Specification (CES) (Ide, 1998). The scheme is based on recognized standards and is therefore likely to be reusable with different software systems. To support our scheme, we propose a data architecture that enables multiple views of a document (based on the CES scheme outlined in Ide [1998]),[3] and a reference linkage system based on Bruneseaux and Romary (1997). These schemes have been developed with an eye toward flexibility and extendibility, in order to be of the widest possible use. In particular, the data architecture enables access to different annotations of a corpus with minimal processing overhead, and allows the simultaneous representation of different (and sometimes incompatible) annotations of the same data. We have tested the scheme by applying it to a small corpus in English, French, and Romanian, and subsequently used it for our research on VT.[4]

In section 2, we describe a tool architecture supporting our encoding scheme. In section 3, we provide a brief overview of VT and demonstrate how the annotated corpora have been used to validate this theory. We then provide an overview of the encoding format in section 4.

## 2. The Annotation Architecture

We have defined a multi-level (hierarchical) parallel annotation architecture compatible with the data architecture defined in the CES that accommodates different annotation views of the same document.. In our scheme, a "hub" document (HD), containing markup for basic document structure down to the level of paragraph as well as (possibly) some sub-paragraph markup for sentence segmentation and/or special tokens such as names, dates, etc., is referenced via inter-document links by a family of documents, each containing an additional view (AD) of the HD.

---

[1] See, for example, the Discourse Resource Initiative at http://www.georgetown.edu/luperfoy/Discourse-Treebank/dri-home.html

[2] XML is the Extended Markup Language, which is likely to become the successor of SGML.

[3] This data architecture has been adopted for a system of corpus-processing tools (LT NSL) available from Edinburgh University; see McKelvie et al. (forthcoming).

[4] Our results using this test data are described in Cristea & Ide (submitted).

The overall architecture is that of a directed acyclic graph (DAG) with the HD as its root, thereby disallowing circular addressing. All documents in the hierarchy represent annotations made from different perspectives of the same original hub document. The markup in all documents along the path from the root to a particular AD are inherited by that AD, i.e., the markup from all parents is combined in the child document.

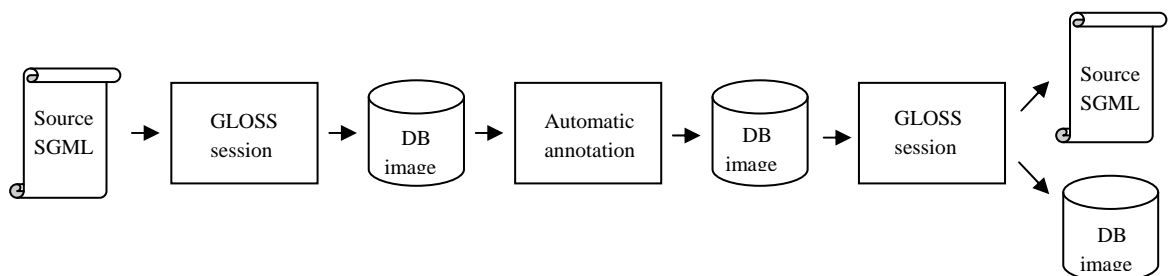

Figure 1: Mixed manual-automatic annotation with GLOSS

To implement this view-based scheme, an annotation tool called GLOSS (Cristea, Craciun and Ursu, 1998), was developed with the following features:
- **SGML compatibility:** the annotator takes as input both plain texts and SGML documents paired with their DTDs[5]. At any point in the annotation process, the document can be saved in SGML format.
- **database image copy**: during the annotation process, an internal representation of the markup is kept in an associated database. When an annotation session is finished, the associated database can be saved for interrogation purposes. Queries addressed to the database can be expressed in SQL[6].
- **manual/automatic annotation**: once a database image of a document exists, it is used as input for a subsequent annotation session with GLOSS. This enables enriching of certain types of tags using an automatic procedure, as outlined in Figure 1.
- **multiple documents/multiple views**: GLOSS allows simultaneously opening more than one document, similar to a text editor. The user can move between documents, each of which has an associated database image. Multiple views are obtained by unifying the database representations of the parent(s) documents. Therefore, when a document inherits from two or more parent documents, another database is generated that copies common parts from these parents and adds the markup that is specific to each of them. Once the parentage relations are established (which occurs when a new view is created), the document loses all connection with its parent documents, such that modifications can be made to the new document without affecting the originals.
- **non-monotonic behavior**: because each document is associated with its own database, the user can perform modifications as follows:
   - creation of a new view defined to inherit from one or more parent views;
   - addition, modification or deletion of attribute-value pairs on elements inherited from parent views, without affecting the view defined by the markup in the ancestor. Added attributes are inherited by hierarchically inferior views defined for the current view;
   - addition of new elements together with their attributes, read-accessible to any inferior view;
   - deletion of inherited elements without affecting the parent view.
- **interactive discourse structure annotation**: annotating the discourse structure in GLOSS is an interactive visual process that aims at creating a binary tree (Marcu, 1996, Cristea and Webber, 1997), where intermediate nodes are relations and terminal nodes are units. Experience gained by authors in manual annotation of discourse structure trees reveals that an incremental, unit-by-unit evolution precisely mimicking an automatic expectation-based parsing (Cristea and Webber, 1997) is not compulsory during a manual process. Manual annotation is closer to a trial-and-error, island-driven process. To facilitate the tree structure building, GLOSS allows development of partial trees that can subsequently be integrated into existing structures by adjoining or substitution.

The principal advantage of this architecture is that it accommodates independent views of the same SGML document. As such, different teams with different expertise can work independently one of the other on the same original document, each accomplishing different annotation tasks. Later, by simply declaring the resulting documents as parent views, GLOSS will combine the different annotations into a single document, retaining only one instantiation of common markup.

## 3. Overview of the Encoding Conventions

The encoding conventions that we adopt for reference annotation and discourse structure are based upon a simple but important principle of separation of segmental and relational markup. *Segmental markup* includes elementary identification of the units of interest for a given study (e.g., referring units, discourse segments, etc.). *Relational markup* identifies structural constraints between these units (e.g., co-referential links, discourse relations, etc.). Separation of these two types of markup has the following advantages:

---

[5] The current implementation allows for a simplified DTD syntax.

[6] The current implementation does not enable database interrogation within the annotator.

- Segmental information is likely to be theory-independent and consensual,[7] whereas the nature and number of relations will change depending on the approach to reference (strict co-referential view, anaphoric chains, etc.).
- Our annotation architecture enables multiple relational encodings for the same segmental level, thus providing potentially several perspectives on the same text.
- Separation of the two types of markup implies two phases in the annotation process of a given document, thus enabling better evaluation of results from each phase.

In our scheme, markup for discourse structure is accomplished using the TEI/CES <SEG> element with attributes *type* (with values such as "unit") and *id* (which provides a unique identifier for each <SEG> element, used for linking). We have added a third attribute, *nuclei,* to enable identification of a nucleus (where appropriate). Relational markup, which identifies structural relationships among segments (e.g., RST relations among discourse units) is encoded using the TEI/CES <LINK> element. <LINKGRP> elements group <LINK> elements that comprise part of the same level of annotation. The overall encoding structure is illustrated by the following:[8]

```
<BODY>
 <DIV>
  <P>
   FIRST UNIT</SEG>
   SECOND UNIT</SEG>
   THIRD UNIT</SEG>
   FOURTH UNIT</SEG>
  </P>
 </DIV>
 <LINKGRP TYPE="RELATION" TARGORDER="Y">
   <LINK ID="L1" SUBTYPE="ELABORATION"
         TARGETS="U1 L2" NUCLEI="U1"/>
   <LINK ID="L2" SUBTYPE="NARRATION"
         TARGETS="L3 U4" NUCLEI="L3 U4"/>
   <LINK ID="L3" SUBTYPE="CIRCUMSTANCE"
         TARGETS="U2 U3" NUCLEI="U3"/>
 </LINKGRP>
</BODY>
```

A similar structure has been adopted for encoding reference, using the <RS> tag to identify segments, as described in detail in Brunesceau and Romary (1997). For example, consider the following fragment [9] (referring expressions are underlined and indexed with IDs for readability):

u1. Il existe quelque chose$^{q1}$ de plus épouvantable que ne l'est l'abandon du père$^{p65}$ par ses deux filles$^{p66}$, qui$^{p67}$ le$^{p68}$ voudraient mort.

u2. C$^{q2}$'est la rivalité$^{q3}$ des deux soeurs$^{p69}$ entre elles$^{p70}$.
u3. Restaud$^{p71}$ a de la naissance,
u4. sa femme$^{p72}$ a été adoptée,
u5. elle$^{p73}$ a été présentée;
u6. mais sa soeur, sa riche soeur, la belle Madame Delphine De Nucingen$^{p74}$, femme d'un homme d'argent$^{p74a}$, meurt de chagrin;
u7. la jalousie la$^{p75}$ dévore,
u8. elle$^{p76}$ est à cent lieues de sa soeur$^{p77}$;
u9. sa soeur$^{p78}$ n'est plus sa soeur$^{p79}$;
u10. ces deux femmes$^{p80}$ se$^{p81}$ renient entre elles$^{p82}$ comme elles$^{p83}$ renient leur père$^{p84}$.

The marked-up version of this fragment is as follows:

```

 
  IL EXISTE QUELQUE CHOSE DE PLUS EPOUVANTABLE QUE NE L'EST L'ABANDON DU
  <RS TYPE="PERSON" ID="P65">PERE</RS> PAR
  <RS TYPE="PERSON" ID="P66">
      SES DEUX FILLES</RS>,
  <RS TYPE="PERSONE" ID="P67">QUI</RS>
  <RS TYPE="PERSON" ID="P68">LE</RS>
   VOUDRAIENT MORT.</SEG>
 C'EST LA RIVALITÉ DES
  <RS TYPE="PERSON" ID="P69">DEUX SOEURS</RS> ENTRE
  <RS TYPE="PERSON" ID="P70">ELLES</RS>.</SEG>
 
  <RS TYPE="PERSON" ID="P71">
   <NAME TYPE="PERSON" KEY="M. DE RESTAUD">
    RESTAUD</NAME></RS> A DE LA NAISSANCE, </SEG>
 
  <RS TYPE="PERSON" ID="P72">SA FEMME</RS>
  A ÉTÉ ADOPTÉE,</SEG>
 
  <RS TYPE="PERSON" ID = "P73">ELLE</RS>
  A ETE PRESENTEE ;</SEG>
 <SEG TYPE = "UNIT" ID="U6"> MAIS
  <RS TYPE="PERSON" ID = "P74">SA SOEUR, SA RICHE SOEUR, LA BELLE
      <NAME TYPE="PERSON" KEY="DELPHINE">
       MADAME DELPHINE DE NUCINGEN</NAME>, FEMME D'
      <RS TYPE="PERSON" ID=P74A>UN HOMME D'ARGENT
      </RS></RS>, MEURT DE CHAGRIN ;</SEG>
  LA JALOUSIE
  <RS TYPE="PERSON" ID="P75">LA</RS> DÉVORE,</SEG>
 
  <RS TYPE = "PERSON" ID="P76">ELLE</RS>
  EST À CENT LIEUES DE
  <RS TYPE="PERSON" ID="P77">SA SOEUR</RS> ;</SEG>
 
   <RS TYPE="PERSON" ID="P78">SA SOEUR</RS>N'EST PLUS
   <RS TYPE="PERSON" ID="P79">SA SOEUR</RS>;</SEG>
 
   <RS TYPE="PERSON" ID="P80">CES DEUX FEMMES</RS>
   <RS TYPE="PERSON" ID="P81">SE</RS> RENIENT ENTRE
   <RS TYPE="PERSON" ID="P82">ELLES</RS> COMME
   <RS TYPE="PERSON" ID="P83">ELLES</RS> RENIENT
   <RS TYPE="PERSON" ID="P84">LEUR PÈRE</RS>.</SEG>
</SEG>
```

---

[7] Well known problems at this level include inclusion of complements in referring units, marking of verb phrases, etc.

[8] For clarity and brevity, the example includes annotations "collapsed" with the Hub Document to form a single SGML document rather than a graph of interrelated documents, as outlined in section 2. However, in reality the different types of markup are included in separate SGML documents.

[9] From Honoré de Balzac, *Le Pere Goriot*

```
<LINKGRP TYPE="COREF PERSON " TARGORDER="Y">
;; PERE GORIOT'S DAUGHTERS
  <LINK TARGETS="P67 P66">
  <LINK TARGETS="P69 P67">
  <LINK TARGETS="P70 P69">
  <LINK TARGETS="P80 P70">
  <LINK TARGETS="P81 P80">
  <LINK TARGETS="P82 P81">
  <LINK TARGETS="P83 P82">
</LINKGRP>

<LINKGRP TYPE="COREF PERSON " TARGORDER="Y">
;; PERE GORIOT
  <LINK TARGETS="P68 P65">
  <LINK TARGETS="P84 P68">
</LINKGRP>

<LINKGRP TYPE="COREF PERSON " TARGORDER="Y">
;; MME DE RESTAUD
  <LINK TARGETS="P77 P72">
  <LINK TARGETS="P78 P77">
  <LINK TARGETS="P79 P78">
</LINKGRP>

<LINKGRP TYPE="COREF PERSON " TARGORDER="Y">
;; MME DE NUCINGEN
  <LINK TARGETS="P75 P74">
  <LINK TARGETS="P76 P75">
</LINKGRP>

<LINKGRP TYPE="RELATION" TARGORDER="Y"
         NUCORDER="Y">    ;; RELATION TYPE LINKS
  <LINK ID="L1" TARGETS="U4 U5" NUCLEI="U4 U5">
  <LINK ID="L2" TARGETS="U3 L1" NUCLEI="U3 L1">
  <LINK ID="L3" TARGETS="U6 U7" NUCLEI="U6 U7">
  <LINK ID="L4" TARGETS="U9 U10" NUCLEI="U9">
  <LINK ID="L5" TARGETS="U8 L4" NUCLEI="U8">
  <LINK ID="L6" TARGETS="L3 L5" NUCLEI="L3">
  <LINK ID="L7" TARGETS="L2 L6" NUCLEI="L2 L6">
  <LINK ID="L8" TARGETS="U2 L7" NUCLEI="U2">
  <LINK ID="L9" TARGETS="U1 L8" NUCLEI="U1">
</LINKGRP>

<LINKGRP TYPE="BRIDGE" TARGORDER="Y">
  <LINK NAME="POSS" TARGETS="P72 P71">
</LINKGRP>
```

As this example shows, we currently base our linkage mechanisms on the TEI extended pointer mechanisms. However, we are exploring the use of the pointer mechanism defined by the WWW Consortium using XML (Maler & DeRose, 1998), which are inspired by the TEI Guidelines and amenable to support by a wide range of software.

## 4. Application of the Architecture to Structure-Reference Study

In Cristea & Ide (1998), we propose an approach to long-distance reference resolution that demonstrates the relation between discourse cohesion and coherence and discourse structure. Our model, which we call *Veins Theory* (VT), is centered on the identification of "veins" over RST-like discourse structure trees. The fundamental assumption underlying VT is that an inter-unit reference is possible *only if the two units are in a structural relation with one another.* In Cristea & Ide (1998) we describe the means by which veins are computed over discourse structure trees and then define domains of accessibility derived from the veins. Accessibility domains for any node in a discourse structure tree may include units which are sequentially distant in the text stream, and thus long-distance references (including those requiring "jumps" over units or segments that contain syntactically feasible referents) can be accounted for. Thus our model provides a description of *global discourse cohesion*, which significantly extends the model of local cohesion provided by Centering Theory (CT) (Grosz, Joshi, and Weinstein 1986, 1995).

The *domain of accessibility* of a unit is defined as the string of units appearing in its vein expression. The main conjecture of VT is that references from a given unit are possible only in its domain of accessibility. Therefore, in VT reference domains for any node may include units that are sequentially distant in the text stream, and thus long-distance references (including those requiring "pops" (Grosz [1977]) over segments that contain syntactically feasible referents) can be accounted for.

A *smoothness score* for a discourse segment can be computed by attaching an elementary score to each transition between sequential units according to Table 2, summing up the scores for each transition in the entire segment, and dividing the result by the number of transitions in the segment. This provides an index of the overall coherence of the segment.

**Table 2: Smoothness scores for transitions**

| | |
|---|---|
| CENTER CONTINUATION | 4 |
| CENTER RETAINING | 3 |
| CENTER SHIFTING (ABRUPT) | 1 |
| CENTER SHIFTING (SMOOTH) | 2 |
| NO Cb | 0 |

A *global CT smoothness score* can be computed by adding up the scores for the sequence of units making up the whole discourse, and dividing the result by the total number of transitions (number of units minus one). In general, this score will be slightly higher than the average of the scores for the individual segments, since accidental transitions at segment boundaries will be included. Analogously, a *global VT smoothness score* can be computed using accessibility domains to determine transitions rather than sequential units. Using this data, we can then compare the smoothness scores using CT and VT.

We claim that the global smoothness score of a discourse when computed following VT is at least as high as the score computed following CT. To validate this claim and VT in general, we implemented the above annotation scheme to encode a small corpus of texts in English, French, and Romanian to use for validating VT. The following texts were included in our analysis:

- three short English texts, RST-analyzed by experts (source: Daniel Marcu) and subsequently annotated for reference and Cf lists by the authors;
- a fragment from Honoré de Balzac's *Le Père Goriot* (French), previously annotated for co-reference (Brunseaux and Romary [1997]); RST and Cf list (see below) annotation made by the authors;

- a fragment from Alexandru Mitru's "Legendele Olimpului"[10] (Romanian); structure, reference, and *Cf* (see below) lists annotated by one of the authors.

As described in section 3, the encoding marks referring expressions, links between referring expressions (co-reference or functional) units, relations between units (if known), and nuclearity. We also include an attribute to encode *forward-looking centers (Cf)* comprising a list of referring expressions, and *backward-looking centers (Cb)*, which consist of a single `<RS>`.[11]

We developed a program[12] that does the following:
- builds the tree structure of units and relations between them;
- adds to each referring expression the index of the unit it occurs in;
- computes the heads and veins for all nodes in the structure;
- determines the accessibility domains of the terminal nodes (units);
- counts the number of direct and indirect references.

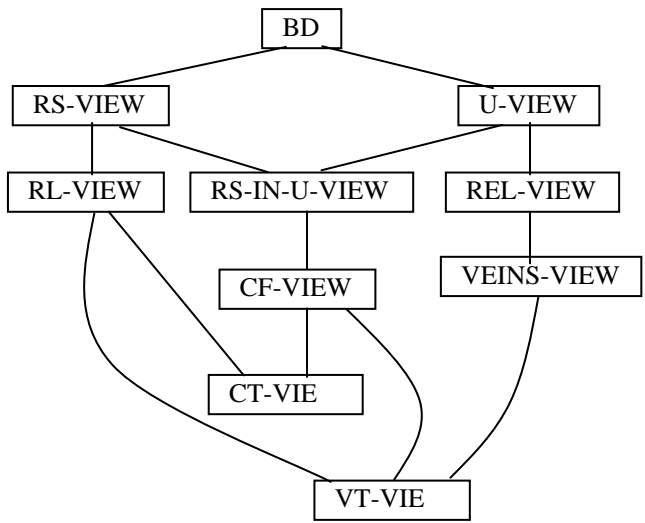

Figure 2: The hierarchy of views for the validation of VT

The hierarchy of views encoded in the documents is given in Figure 2. The views include:

- **BD:** the base document, containing the unannotated text and possibly markup for basic document structure down to the level of paragraph.
- **RS-VIEW:** includes markup for isolated reference strings. The basic elements are RS (reference strings).
- **RL-VIEW:** the reference links view, imposed over the RS-VIEW, includes reference links between an anaphor, or source, and a referee, or target. Links configure co-reference chains, but can also indicate bridge references (Hahn and Strübe, 1997; Passoneau, 1994, 1996).
- **U-VIEW:** marks discourse units (sentences, and possibly clauses). Units are marked as `<seg>` elements with *type=unit*.
- **REL-VIEW:** reflects the discourse structure in terms of a tree-like representation.
- **VEINS-VIEW:** includes markup for head and vein expressions. *Head* and *vein* attributes (with values comprising lists) are added to all `` and `<LINK TYPE=RELATION>` elements.
- **RS-IN-U-VIEW:** inherits `<RS>` and `` elements from U-VIEW and RS-VIEW. It also includes markup that identifies the discourse unit to which a referring string belongs.
- **CF-VIEW:** inherits all markup from RS-IN-U-VIEW, and adds a list of forward looking centers (the *cf* attribute) to each unit in the discourse.
- **CT-VIEW** *(Centering Theory view)***:** inherits CF lists from the CF-VIEW and backward references from the RL-VIEW. Using the markup in this view, transitions can be computed following classical CT.
- **VT-VIEW** *(Veins Theory view)*: inherits CF centers from the CF-VIEW, back-references from the RL-VIEW, and vein expressions from the VEINS-VIEW. The VT-VIEW also includes markup for Cb's computed along the veins of the discourse structure (the $cb\text{-}h$[13] attribute of the `` elements).

Once the documents were encoded, Cb's and transitions were determined following the sequential order of the units (according to classical CT), and a smoothness score was computed. Then, following VT, accessibility domains were used to determine maximal chains of accessibility strings, Cb's and transitions were determined based on these strings, and a VT smoothness score was similarly computed. The results are summarized in Table 1, which shows that the score for VT is better than that for CT in all cases.

### 5. Conclusion

In this paper we outline an encoding scheme and a data architecture for discourse, together with a set of tools that support the annotation of corpora. We have used these tools to annotate corpora in English, French, and Romanian and used them to study a model of discourse cohesion based on Veins Theory. Our results demonstrate that VT provides a promising approach to identifying domains of referential accessibility in discourse.

There is, at present, no encoding standard for discourse. The few annotated corpora available are encoded using a variety of formats, which in turns often demands re-encoding when these corpora are used with different pieces of software. In our view, it is essential to not only determine a standard for encoding discourse, but also to define a data architecture which is maximally flexible. The view-based architecture and inheritance mechanism described in this paper provide a viable framework for discourse encoding, which allows the representation of a variety of types of annotation and can accommodate different theories and perspectives. We are currently exploring the extension of our scheme to support multi-lingual analyses; this should be readily accomplished using linkage mechanisms similar to those described here to associate parallel text passages.

---

[10] "The Legends of Olimp"

[11] In CT each unit is associated with a list of forward-looking centers (*Cf lists)*, where elements are partially ranked according to discourse salience; and a unique backward-looking center (*Cb),* that is the first center in the Cf list of the previous unit also realized in the current unit.

[12] Written in Perl and run on a Sun workstation.

[13] From "hierarchical".

| Source | No. of transitions | CT Score | Average CT score per transition | VT score | Average VT score per transition |
|---|---|---|---|---|---|
| English | 59 | 76 | 1.25 | 84 | 1.38 |
| French | 47 | 109 | 2.32 | 116 | 2.47 |
| Romanian | 65 | 142 | 2.18 | 152 | 2.34 |
| **Total** | **173** | **327** | **1.89** | **352** | **2.03** |

Table 1: CT smoothness scores vs. VT smoothness scores